\documentclass[letterpaper, 10 pt, conference]{ieeeconf}
\IEEEoverridecommandlockouts
\usepackage{cite}
\usepackage{amsmath,amssymb,amsfonts}
\usepackage{algorithmic}
\usepackage{graphicx}
\usepackage{textcomp}
\usepackage{float}
\usepackage{mathtools}
\usepackage{subcaption}
\usepackage{amsmath}
\usepackage{xcolor}

\def\BibTeX{{\rm B\kern-.05em{\sc i\kern-.025em b}\kern-.08em
    T\kern-.1667em\lower.7ex\hbox{E}\kern-.125emX}}
\begin{document}


\title{Adaptive Gravity Compensation Control of a Cable-Driven\\ Upper-Arm Soft Exosuit}

\IEEEoverridecommandlockouts 

\author{Joyjit Mukherjee$^{1}$, Ankit Chatterjee$^{2}$, Shreeshan Jena$^{3}$, Nitesh Kumar$^{4}$,\\ Suriya Prakash Muthukrishnan$^{5}$, Sitikantha Roy$^{3}$, Shubhendu Bhasin$^{3 \ast}$
\thanks{$^{\ast}$S. Bhasin is the corresponding author.}%
\thanks{This research is supported by DRDO-JATC project
number RP04191G.}%
\thanks{$^{1}$J. Mukherjee is with Birla Institute of Technology and Science Pilani, Hyderabad Campus, Hyderabad, India.
        {\tt\footnotesize j.mukherjee@hyderabad.bits-pilani.ac.in}}%
\thanks{$^{2}$A. Chatterjee is with PricewaterhouseCoopers, India.
        {\tt\footnotesize chatterjee.ankit835@gmail.com}}%
\thanks{$^{3}$S. Jena, N. Kumar, S. Roy, and S. Bhasin are with the Indian Institute of Technology Delhi, India.
        {\tt\footnotesize shreeshan24@gmail.com, nitesh@alumni.iitm.ac.in, sroy@am.iitd.ac.in, sbhasin@ee.iitd.ac.in}}%
\thanks{$^{5}$S. P. Muthukrishnan is with the All India Institute of Medical Sciences, New Delhi, India.
        {\tt\footnotesize dr.suriyaprakash@aiims.edu}}%
}

\maketitle

{\begin{abstract}
This paper proposes an adaptive gravity compensation (AGC) control strategy for a cable-driven upper-limb exosuit intended to assist the wearer with lifting tasks. Unlike most model-based control techniques used for this human-robot interaction task, the proposed control design does not assume knowledge of the anthropometric parameters of the wearer's arm and the payload. Instead, the uncertainties in human arm parameters, such as mass, length, and payload, are estimated online using an indirect adaptive control law that compensates for the gravity moment about the elbow joint. Additionally, the AGC controller is agnostic to the desired joint trajectory followed by the human arm. For the purpose of controller design, the human arm is modeled using a 1-DOF manipulator model. Further, a cable-driven actuator model is proposed that maps the assistive elbow torque to the actuator torque. The performance of the proposed method is verified through a co-simulation, wherein the control input realized in MATLAB is applied to the human bio-mechanical model in OpenSim under varying payload conditions. Significant reductions in human effort in terms of human muscle torque and metabolic cost are observed with the proposed control strategy. Further, simulation results show that the performance of the AGC controller converges to that of the gravity compensation (GC) controller, demonstrating the efficacy of AGC-based online parameter learning. 
\end{abstract}}


\section{Introduction}

Wearable assistive devices are a rapidly emerging field of study aimed at augmenting human abilities or restoring lost function. These wearable devices can vary from simple braces and supports to more advanced robotic exoskeletons and prostheses. Wearable assistive technologies have the potential to enhance human strength \cite{lee2022recent}, reduce metabolic cost in activities of daily living (ADLs) \cite{Collins,quinlivan2017assistance}, improve the quality of life for people with spinal cord injuries, stroke, or neuromuscular illnesses \cite{Westlake,Frisoli,Tsagarakis, Esquenazi }, and assist in rehabilitation \cite{Tsagarakis}. Wearable assistive devices are getting more sophisticated, lighter, and inexpensive as technology advances, making them more accessible worldwide.

An assistive device can either be active or passive \cite{Ollinger}. A passive assistive device does not provide additional energy to the user. It is designed to assist the user's movements by improving the efficiency of their own muscular power. Examples of passive assistive devices include orthotics, braces, and other support devices that can help people with physical disabilities move more easily and with less pain or discomfort. Active devices, on the other hand, behave as energy sources and provide aid proportional to human intention. Intention detection can be achieved by interfacing with the human nervous system to extract neural signals, specifically by using electromyography (EMG) activity to estimate muscle activation. Model-based myoelectric controllers \cite{Lotti} and EMG-driven machine learning controllers \cite{Lotti1} have been used to deliver assistance that is proportionate to the wearer's muscular engagement. Since EMG is generated prior to movement initiation, it is often used as a feedforward term to align with the user's movement timing. Despite the theoretical benefits of EMG-based approaches, empirical data is scarce on how users behave toward various intention detection methodologies implemented on identical devices.

Mechanically intrinsic methods of intention detection are complementary to EMG-based control since they rely on mechanical representations of human motions gathered by sensors on the device itself \cite{Lotti2}. These data are then used with the inverse dynamics model of the supported limb to compute motion kinetics, which are then leveraged in control design. Mechanically intrinsic controllers are highly researched and durable, however, as shown in \cite{Lotti2}, these strategies have to deal with the time delay between the intention detection and actuation, and are dependent on the exact knowledge of the system parameters, such as limb mass, length, payload, etc.

A popular mechanically intrinsic approach is gravity compensation (GC) control \cite{Domenico,Georgarakis}, which relies on the knowledge of system parameters and measurement of joint angles to compensate for the gravity moment, and thus, reduce human effort. GC control strategy is simple to implement and requires minimal use of sensors; however, it is user-specific and has limited adaptability to dynamic load conditions.

To address these limitations, this paper presents an adaptive gravity compensation (AGC) control method that does not assume knowledge of the system parameters and is adaptable to dynamic changes in the system and environment interactions, thus providing greater flexibility in the control of wearable robotic devices. Unlike EMG-based controllers that require a lengthy calibration and training phase \cite{Lotti3}, the AGC method does not require any offline phase and automatically adjusts, in real-time, to uncertainty in the system parameters. Different from the classical adaptive control methods \cite{narendra2012stable}, the additional challenge in designing an adaptive controller for this human-robot interaction problem is the unavailability of the reference/desired human trajectory, which is used to compute the trajectory tracking error and automatic controller tuning. In the absence of tracking error information, the proposed controller uses the prediction error \cite{slotine1989composite} to adaptively learn the system parameters online.  

The AGC controller achieves compensation of the gravitational forces \cite{Dinh} acting on the elbow through a cable-driven actuator connected to the exosuit. The human arm is modeled as a 1-DOF manipulator, and the tendon-sheath actuator considered here is based on Hill’s muscle model \cite{Wolbrecht,Miller,Zhang}. A cable-driven actuator model is proposed in this work that maps the assistive torque at the elbow to the torque applied by the actuator. The control system is implemented using MATLAB and OpenSim, an open-source biomechanical simulation platform. The simulation results presented in this work are aimed at assessing the feasibility of using this approach for human augmentation.

\section{System Modeling} \label{sec1}

The assistance of the upper limb is challenging due to the dynamic nature of human tasks, ranging from fine manipulation and force compensation to general interaction with the external environment having unpredictable dynamics \cite{Gui}. The requirement of symbiotic synergy with the human brain and non-planar motion of the upper limb adds to the complexity and complicates the biomechanical analysis. In this work, we have adopted a simplified approach where motion is considered in the sagittal plane, and a $1$-DOF model of the elbow flexion-extension is considered.

\begin{figure}[t]
\centering
\includegraphics[width=0.95\linewidth]{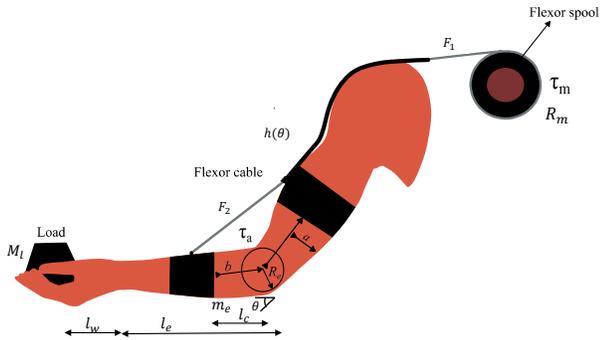}
\caption{Cable-Driven Actuator on Human Hand.}
\label{fig:1}
\end{figure}

\begin{table}[h]
  \begin{center}
    \caption{Nomenclature}
    \label{tab:table1}
    \begin{tabular}{l||l} 
    \hline
      \textbf{Symbol} & \textbf{Description}\\
      \hline
      $m_e$ & Mass of the forearm\\
      $M$ & Additional mass\\
      $l_e$ & Length of the forearm\\
      $b_e$ & Damping coefficient at elbow joint\\ 
      $I_e$ & Moment of Inertia of forearm\\
      $l_c$ & Distance between elbow joint and Forearm CG\\ 
      $l_w$ & Distance between end of forearm to the centre of wrist\\ 
      $a$ & Half Width of forearm at strap\\
      $b$ & Distance between elbow joint and strap\\
      $R_m$ & Radius of spool\\
      $\theta_m$ & Motor angle\\
      $\mu$ & Friction coefficient between cable and sheath\\
      $\phi$ & Curvature of sheath\\
      $\theta$ & Angular position of elbow Joint\\ 
      $\tau_h$ & Torque generated by human hand\\
      $\tau_a$ & Assistive torque provided at elbow joint\\
      $\tau_m$ & Motor torque providing assistance \\
      $N$ & Gear Ratio \\
\hline
    \end{tabular}
  \end{center}
\end{table}

\subsection{Upper Limb Model}
\begin{figure*}[ht]
\centering
\includegraphics[width=0.9\linewidth]{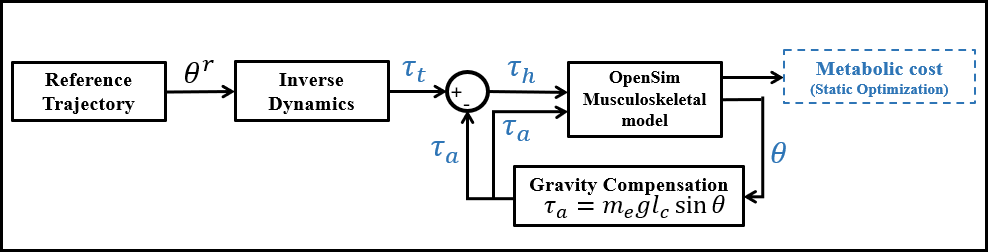}
\caption{ Block Diagram of Gravity Compensation (GC) Control Framework.}
\label{fig:3}
\end{figure*}

The dynamic equations of motion for the human forearm, as shown in Figure \ref{fig:3}, with actuation provided at the elbow joint can be expressed as
\begin{equation} \label{sys}
   I_e \ddot{\theta} + b_e \dot{\theta} + \{m_e l_c + M (l_e + l_w) \} g \sin (\theta) = \tau_h + \tau_{a}
\end{equation}
The parameters used to model the human forearm are given in Table \ref{tab:table1}. The dynamic equation is modeled as a $1$-DOF manipulator with actuation at the elbow joint. The actuation is provided by two sources, the human muscles generating a moment about the elbow joint $\tau_h(t)$ and actuation mechanisms in the exosuit $\tau_a(t)$. The exosuit is designed such that the lines of action of the forces coincide with that of the human muscles to achieve optimum synergy, which at the same time assures that any natural human motion is not inhibited. The exosuit considered in this work is actuated by a cable-driven actuator \cite{Miller},\cite{Zhang} described in the next subsection.


\begin{figure*}[t]
\centering
\includegraphics[width=0.9\linewidth]{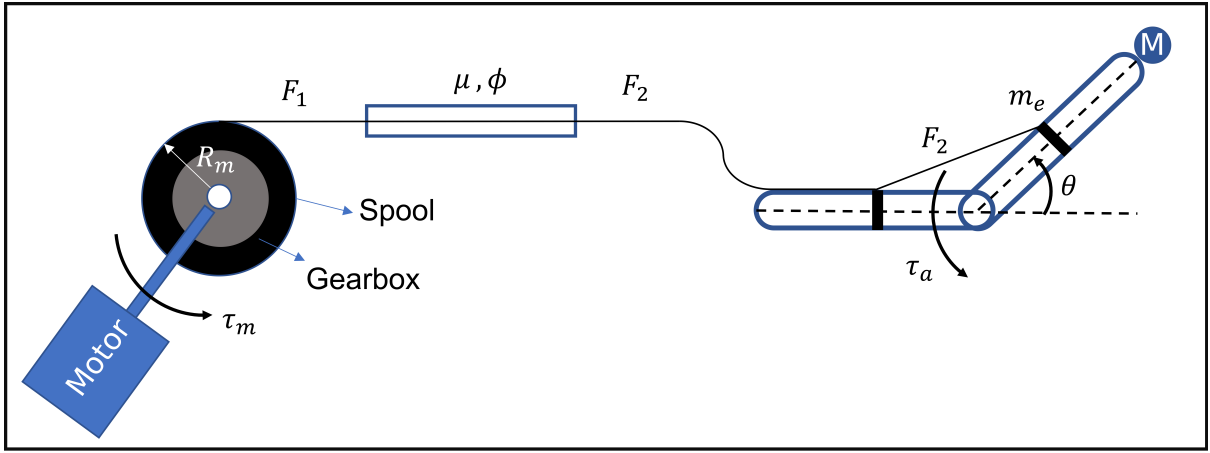}
\caption{Schematic of the Power Transmission Mechanism}
\label{fig:2}
\end{figure*}

\subsection{Cable-Driven Actuator Model}
The cable-driven actuator for actuating the exosuit consists of a single cable, driven by a motor, to generate moment at the elbow joint. A schematic of the cable arrangement, along with all the tensions and moments, is shown in Figure \ref{fig:2}. Conventional cable-driven actuators have a dual-cable arrangement analogous to the agonist-antagonist muscle pairs - one for flexion and the other for extension \cite{Dinh}. This arrangement is suitable for applications like rehabilitation or therapy where assistance in both directions might be required. However, for the scenario considered in this work, i.e., augmentation, it is assumed that the wearer has significant motor skills to perform activities of daily living (ADLs)  but requires strength augmentation for tasks such as lifting heavy loads. Here, the flexor cable does the "heavy lifting" by providing assistance against gravity.  Therefore, the extensor cable, which is supposed to provide assistance during the extension motion,  is not considered in this work. When the motor is actuated, it applies tension to the cable at an appropriate time, depending on the desired flexion motion. The cable passes through guiding sheaths stretching over the back and the shoulder and ending at a strap fixed firmly on the upper arm. The cable continues beyond this without any guiding sheath up to another strap fixed firmly on the forearm and is free to move between the straps, allowing it to create a moment through tension on the elbow joint. Considering $\tau_m$ as the motor torque with $N$ as the gear ratio and $R_m$ as the radius of the spindle attached to the gearbox and the cable, the force in the cable just beyond the spindle can be modeled as
\begin{align}\label{Tension_1} 
F_1 &= \frac{N\tau_m}{R_m}
\end{align}
The cable moves through the guided sheath to transfer optimum power along the axis coincidental with the upper-arm muscle. The relative motion between the cable and the sheath results in a frictional loss and the direction of the frictional force depends on the direction of motion. Therefore, the cable tension beyond the sheath can be expressed as 
\begin{align}\label{Tension_2} 
F_2 &= \begin{cases}
F_1 e^{-\mu\phi},& \text{if } \dot\theta_m\geq0 \\
F_1 e^{\mu\phi},& \text{if } \dot\theta_m<0 \\
\end{cases}
\end{align}
where $\mu_1$ is the coefficient of friction between the cable and the sheath and $\phi$ is the curvature angle of the cable sheath.

The parameter $h (\theta)$ denotes the extension function for the cable. The extension function of the cable $h (\theta)$ represents the variation in length in the cable as a function of the elbow angle written as
\label{Tension_4}
\begin{align}
h (\theta) &= 2 \sqrt{a^2 + b^2} \cos \Big \{ \tan^{-1} \big(\frac{a}{b} \big) + \frac{\theta}{2} \Big \} - 2b
\end{align}
The total assistive torque at the elbow joint due to tension in the cable is given by
\begin{align} \label{Tension_5}
\tau_a = \frac{\partial h}{\partial \theta}F_2
\end{align}
Combining \eqref{Tension_1}, \eqref{Tension_2} and \eqref{Tension_5}, the actuator torque can be mapped to the motor torque as
\begin{align}\label{Tension_6}
\tau_a = G \tau_m,
\end{align}
where
\begin{align} \label{G_transmission}
G &=\begin{cases} 
\frac{NJ_f}{R_m} e^{-\mu\phi},& \text{if } \dot\theta_m\geq0 \\
\\
\frac{NJ_f}{R_m} e^{\mu\phi},& \text{if } \dot\theta_m<0 \\
\end{cases},
\end{align}
where, $J_f = \frac{\partial h}{\partial \theta}$ represents  the moment-arm about the elbow joint. The actuator torque $\tau_a$ will be applied using the cable-driven actuator at the elbow joint to provide the desired assistance. Employing the actuator model \eqref{Tension_6} in the system dynamics \eqref{sys} yields
\begin{equation} \label{sys_1}
 I_e \ddot{\theta} + b_e \dot{\theta} + \{m_e l_c + M (l_e + l_w) \} g \sin (\theta) = \tau_h + G \tau_{m}
\end{equation}

\begin{figure*}[t]
\centering
\includegraphics[width=0.9\linewidth]{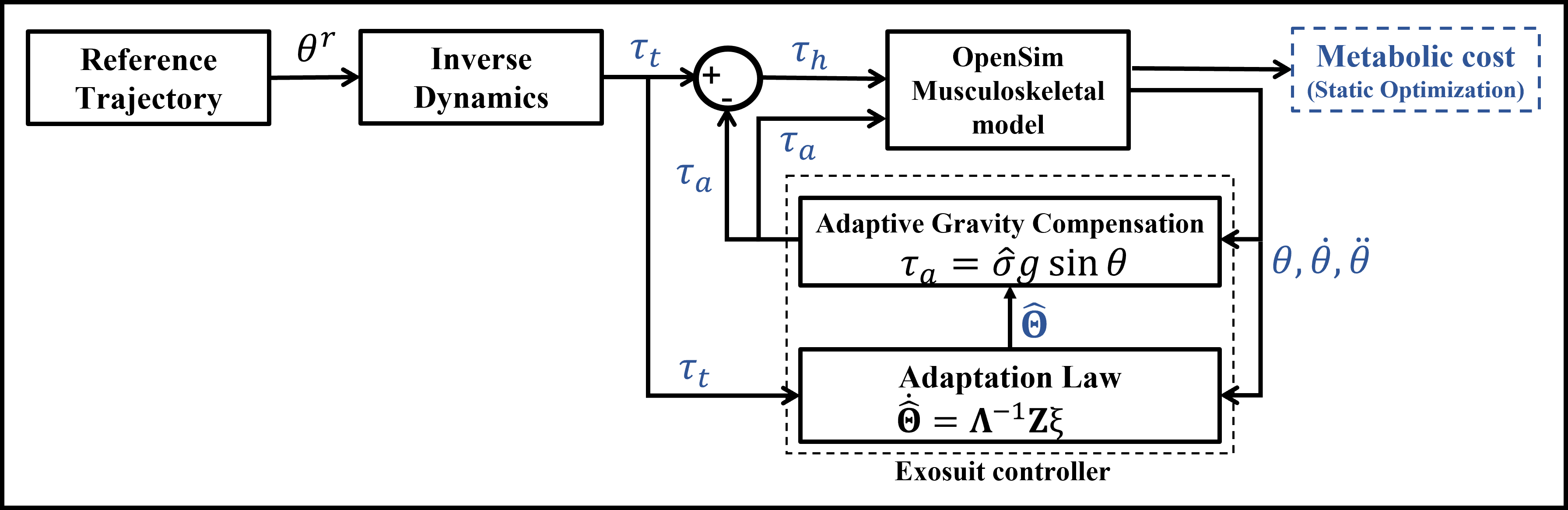}
\caption{Block Diagram of Adaptive Gravity Compensation (AGC) Control Framework.}
\label{fig:4}
\end{figure*}

\subsection{Metabolic Cost Computation}


A simulation platform, demonstrated in our previously published work \cite{sambhav2022integrated}, is utilized in this study to compute the joint moment and metabolic cost parameters. This musculoskeletal simulation platform OpenSim is used to examine the effects of different external load conditions and actuator assistance on these parameters. The metabolic cost of the activity is calculated by combining the rate of heat liberated from the body and the rate of work done as 

\begin{align}\label{mct}
\dot{E}&=\dot{B}+\dot{A}+\dot{M}+\dot{S}+\dot{W}
\end{align}
where, $\dot{B}$ is the basal heat rate, $\dot{A}$  is the activation heat rate, $\dot{M}$ is the maintenance heat rate, $\dot{S}$  is the shortening heat rate and $\dot{W}$ is the mechanical work rate $(W)$.

\section{Control Methodology} \label{sec2}
A tracking controller typically requires knowledge of the desired trajectory for feedback control. However, in the case of assistance, the desired trajectory is generated by the human and is mostly unknown to the controller. Although considerable progress has been achieved in the estimation of human intention from brain and muscle signals, satisfactory results that can be employed on a working exosuit are still elusive. This section first details the design of a gravity compensation (GC) control approach, using which the exosuit compensates for the gravity terms in the arm dynamics and provides a gravity-free environment for the hand to execute the desired motion. This approach is expected to reduce metabolic costs and result in improved endurance. Later this approach is extended to a situation where the hand may be required to lift an unknown payload during flexion. To implement an efficient gravity compensation in such a situation, it is essential to estimate the gravity terms in the dynamic equations and update the control law based on the estimates. This proposed approach has been termed as Adaptive Gravity Compensation (AGC) control. This work primarily focuses on enhancing existing human abilities through an exosuit, although they can also be utilized for rehabilitation in case of reduced motion capabilities. Therefore, the human is assumed capable enough to follow a trajectory based on his own volition and the assistive device works to reduce the human effort in achieving this objective. 

\subsection{Gravity Compensation (GC) Control}
The block diagram of the proposed GC control scheme is given in Figure \ref{fig:3}. Considering all the parameters in the system \eqref{sys_1} to be known, the control law for the motor torque can be written as
\begin{equation} \label{gc_law}
\tau_m = G^{-1} \{m_e l_c + M (l_e + l_w) \} g \sin (\theta)
\end{equation}
where $G$ is defined in \eqref{Tension_6}. The forearm system controlled by the human brain can be obtained by substituting the GC control law \eqref{gc_law} in the arm dynamics \eqref{sys_1} as
\begin{equation} \label{sys_2}
I_{e} \ddot{\theta} + b_{e} \dot{\theta} = \tau_{h}
\end{equation}
A simple feedback control law based on computed torque control can be designed to imitate the tracking response of the human brain as
\begin{align} \label{fb_law}
    \tau_h = I_{e} \ddot{\theta}^r + b_{e} \dot{\theta} - I_e \{ k_p e + k_d \dot{e} \},
\end{align}
where $\theta^r(t)$ represents the desired trajectory with the tracking error being represented as $e(t) \triangleq \theta(t) - \theta^r(t)$.
The control gains $k_p, k_d > 0$ are chosen appropriately to achieve the desired tracking performance. Utilizing the feedback law \eqref{fb_law} in \eqref{sys_2}, the closed loop system can be written as
\begin{align} \label{clsd_sys}
    I_{e} \{ \ddot{e} + k_d \dot{e} + k_p e \} = 0
\end{align}
The closed loop dynamics in \eqref{clsd_sys} results in asymptotic tracking of the desired trajectory, i.e. $e(t) \to 0$ as $t \to 0$.

\subsection{Adaptive Gravity Compensation (AGC) Control} \label{sec3}
The GC control law in \eqref{gc_law} assumes knowledge of the system parameters, i.e. mass, length, the inertia of the forearm, payload, etc. that limits its application in real-world situations where this information may not be available. To address this limitation, an AGC control strategy is proposed. 
The gravity terms can be combined together and expressed as
\begin{align} \label{sys_dyn_1}
&I_e \ddot{\theta} + b_e \dot{\theta} + \{m_e l_c + M (l_e+l_w) \} g \sin (\theta) = \tau_h + G \tau_m, \nonumber \\
&I_e \ddot{\theta} + b_e \dot{\theta} + Y \sigma = \tau_h + G \tau_m,
\end{align} 
where $Y \triangleq g \sin(\theta)$ and $\sigma \triangleq m_e l_c + M (l_e+l_w)$ represents the unknown parameter.
The assistive control law for the motor torque can be expressed as
\begin{align} \label{grav_cont}
\tau_m = G^{-1} Y \hat{\sigma},
\end{align}
where $\hat{\sigma} (t)$ represents the estimated value of the unknown parameter $\sigma$.
Using the control law \eqref{grav_cont} in system \eqref{sys_dyn_1} yields
\begin{align} \label{sys_2}
I_e \ddot{\theta} + b_e \dot{\theta} + Y \tilde{\sigma} = \tau_h,
\end{align}
where $\tilde{\sigma} = \sigma - \hat{\sigma}$ represents the estimation error. Substituting the torque generated by the human brain $\tau_h$ \eqref{fb_law} in \eqref{sys_2}. 
\begin{align} \label{clsd_sys}
    I_{e} \{ \ddot{e} + k_d \dot{e} + k_p e \} = -Y \tilde{\sigma}
\end{align}
\subsection{Adaptation Law}

To estimate the unknown parameter representing the moment due to the mass of the hand and the unknown load on the elbow joint, a prediction error for parameter estimation is defined. The dynamic equation \eqref{sys_dyn_1} can be expressed as
\begin{align} \label{sys_dyn_2}
I_e \ddot{\theta} + b_e \dot{\theta} + Y \sigma = \mathbf{Z} \boldsymbol{\Theta} = \tau_h + G \tau_m,
\end{align} 
where
\begin{align*}
\mathbf{Z} \triangleq \begin{bmatrix}
\ddot{\theta} & \dot{\theta} & Y
\end{bmatrix}, \boldsymbol{\Theta} \triangleq \begin{bmatrix}
I_e & b_e & \sigma
\end{bmatrix}^T
\end{align*}
The vector $\mathbf{Z}(t)$ can be computed from the measurements, however, the actual values of $\boldsymbol{\Theta}$ are not available. Assuming the hand torque $\tau_h(t)$ is measurable, the right hand side of \eqref{sys_dyn_2} and the AGC controller in \eqref{grav_cont} can be used to determine the prediction error as
\begin{align} 
\xi & \triangleq \mathbf{Z} \boldsymbol{\Theta} - \mathbf{Z} \hat{\boldsymbol{\Theta}} = \mathbf{Z} \tilde{\boldsymbol{\Theta}}, \label{pred} \\
\xi &= \tau_h + G \tau_m - \mathbf{Z} \hat{\boldsymbol{\Theta}},
\end{align}
where $\hat{\boldsymbol{\Theta}}(t)$ represents the estimated physical parameters of the system and $\tilde{\boldsymbol{\Theta}} \triangleq \boldsymbol{\Theta} - \hat{\boldsymbol{\Theta}}$ denotes the parameter estimation error.
The parameter estimate $\hat{\sigma}(t)$ is extracted from $\hat{\boldsymbol{\Theta}}(t)$ to implement the assistive control law \eqref{grav_cont}.
A block diagram of the proposed AGC control scheme is given in Figure \ref{fig:4}.

\subsection{Stability Analysis}
Considering a Lyapunov function candidate
\begin{align} \label{lyap}
V = \frac{1}{2} \tilde{\boldsymbol{\Theta}}^T \boldsymbol{\Lambda} \tilde{\boldsymbol{\Theta}},
\end{align}
where $\boldsymbol{\Lambda}$ is positive definite. Taking time-derivative of \eqref{lyap} and considering the unknown load to be constant, i.e. $\dot{\boldsymbol{\Theta}} = 0$ yields
\begin{align} \label{lyap_1}
\dot{V} = - \tilde{\boldsymbol{\Theta}}^T \boldsymbol{\Lambda} \dot{\hat{\boldsymbol{\Theta}}}
\end{align}
The adaptation law for updating the estimated parameters is designed as
\begin{align} \label{adap}
\dot{\hat{\boldsymbol{\Theta}}} = \boldsymbol{\Lambda}^{-1} \mathbf{Z}^T \xi
\end{align}
Employing the adaptation law \eqref{adap} in \eqref{lyap_1} yields
\begin{align} \label{lyap_2}
\dot{V} = - \tilde{\boldsymbol{\Theta}}^T \boldsymbol{\Lambda} \boldsymbol{\Lambda}^{-1} \mathbf{Z}^T \xi
\end{align}
Employing \eqref{pred}, one can write
\begin{align} \label{lyap_3}
\dot{V} = -  \tilde{\boldsymbol{\Theta}}^T \mathbf{Z}^T \mathbf{Z}  \tilde{\boldsymbol{\Theta}} \leq 0,
\end{align}
since $\mathbf{Z}^T \mathbf{Z} \geq 0$ is positive semi-definite. This proves that the estimation error $\tilde{\boldsymbol{\Theta}}(t)$ (and hence, $\tilde{\sigma}(t)$) is bounded. Parameter convergence is not guaranteed unless certain restrictive conditions of persistence of excitation \cite{narendra2012stable} are met. However, boundedness of $\tilde{\sigma}(t)$ ensures that the tracking error $e(t)$ is also bounded, from the stable linear dynamics in \eqref{clsd_sys}. As described earlier, to realise the AGC controller \eqref{grav_cont}, the estimated parameter $\hat{\sigma}$ can be obtained as
\begin{align} \label{adap}
\hat{\sigma} = \begin{bmatrix}
0 & 0 & 1
\end{bmatrix} \hat{\boldsymbol{\Theta}}
\end{align}

\begin{figure}[h]
\centering
\includegraphics[width=0.55\linewidth]{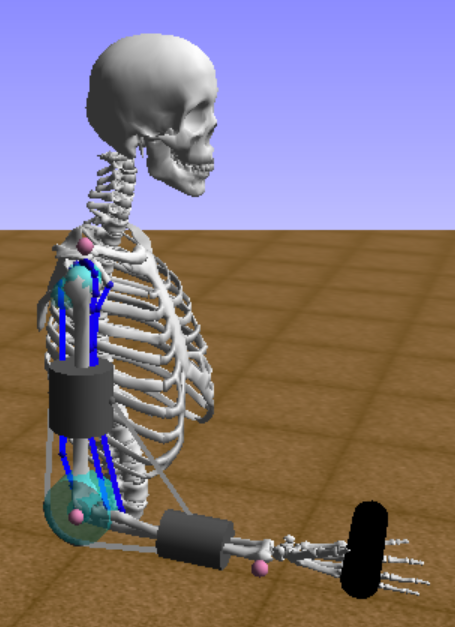}
\caption{Elbow Flexion in OpenSim with the arm26 musculoskeletal model.}
\label{fig:5}
\end{figure}

\section{Simulation Results} \label{sec4}

Simulation studies were conducted to analyse and validate the performance of the proposed approach in a MATLAB - OpenSim co-simulation environment. 
OpenSim is an open-source software for biomechanical modeling, simulation, and analysis \cite{delp2007opensim}. An arm26 musculoskeletal model for the human arm, shown in Figure \ref{fig:5}, with two degrees of freedom \footnote{Only the elbow degree of freedom is active here.} and actuated by six muscles, is used. The metabolic cost is obtained 
from OpenSim using \eqref{mct}. The cable-driven actuator augments the human arm, which includes an additional payload in the hand, as shown in Figure \ref{fig:5}.  The performance of the GC and AGC-based control is compared with the unassisted case. The values of the system parameters considered for simulation are given in Table \ref{tab:table2}.

\begin{table}[h!]
  \begin{center}
    \caption{System parameters}
    \label{tab:table2}
    \begin{tabular}{l||l} 
    \hline
      \textbf{Parameters} & \textbf{Value}\\
      \hline
      $m_e$ & $1.5343$ kg\\
      $l_e$ & $0.28$ m\\
      $b_e$ & $0$ N-s/rad\\ 
      $I_e$ & $0.0201$ kg-m$^{2}$\\
      $l_c$ & $0.1815$ m\\ 
      $l_w$ & $0.04$ m\\ 
      $a$ & $0.05$ m\\
      $b$ & $0.1$ m \\
      $R_m$ & $0.03$ m\\
      $\mu$ & $0.07$ \\
      $\phi$ & $\pi$ rad\\
      $N$ & $25$\\
\hline
    \end{tabular}
  \end{center}
\end{table}

 \begin{figure*}[h!] 
	  \begin{subfigure}{1.1\columnwidth}
	  \flushleft
      \includegraphics[width = 3.6in]{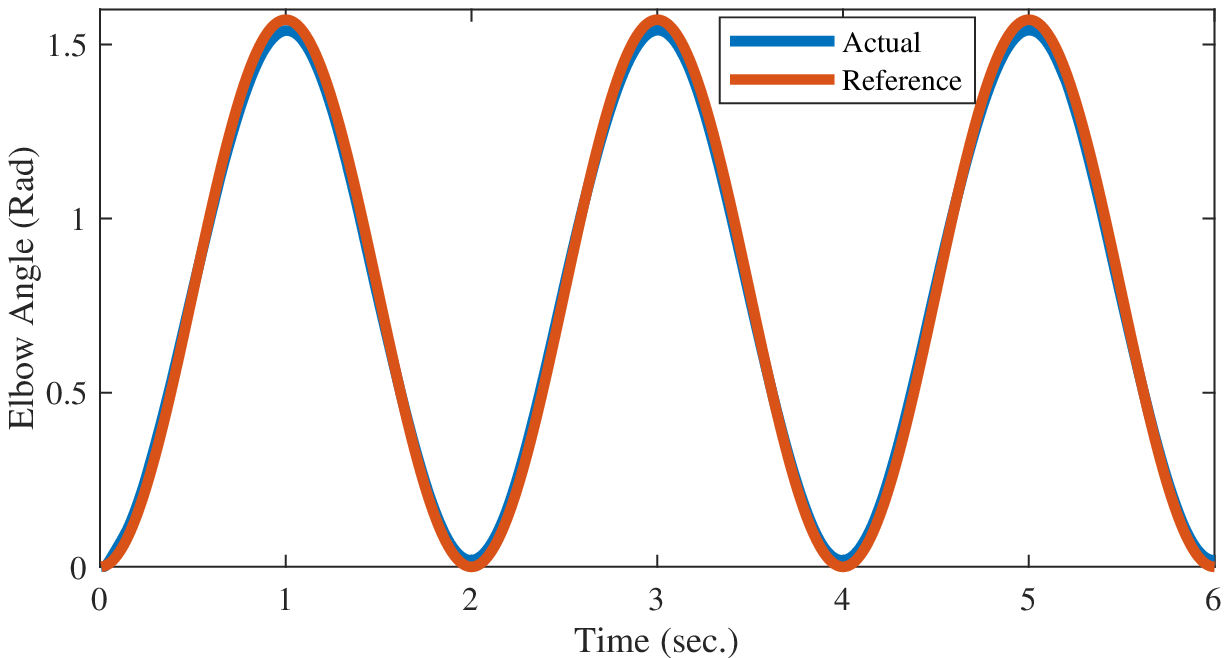}
  \caption{Actual and reference Elbow Trajectories}
      \label{traj_1}  
	  \end{subfigure}  \begin{subfigure}{1.1\columnwidth}
	  \flushleft
      \includegraphics[width = 3.6in]{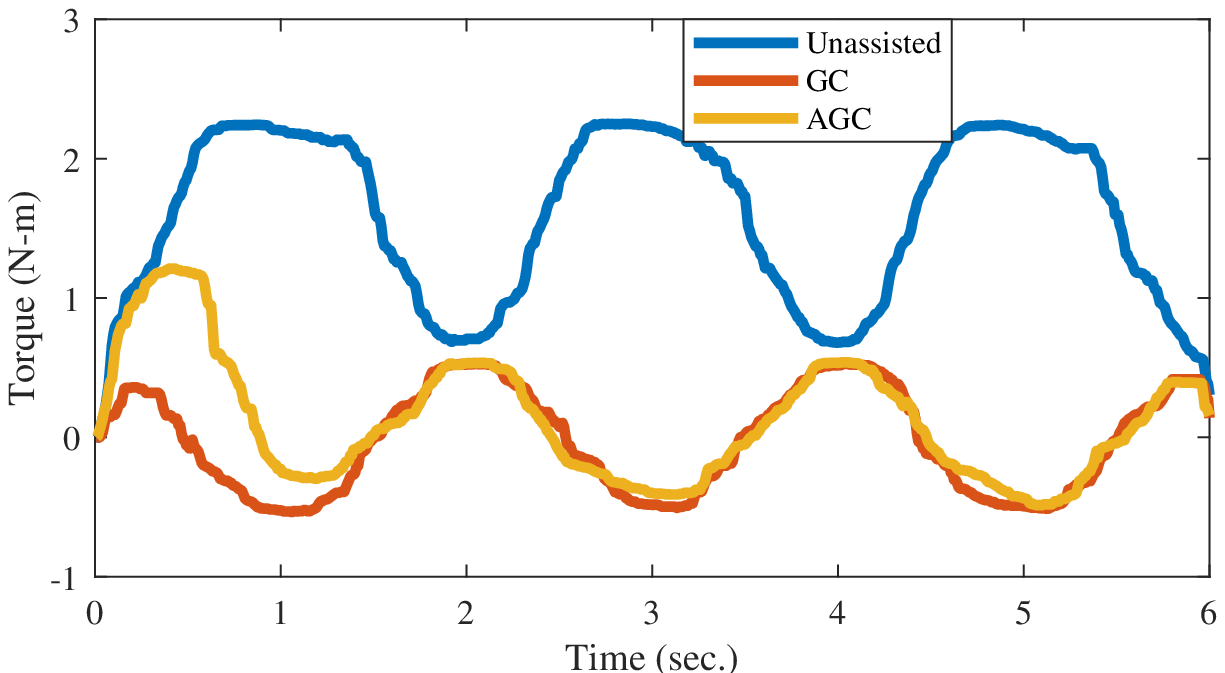}
      \caption{Human Muscle Torque $\tau_h(t)$ at the Elbow}
      \label{Human_torque}  
	  \end{subfigure}  
      \begin{subfigure}{1.1\columnwidth}
	 \centering
      \includegraphics[width = 3.6in]{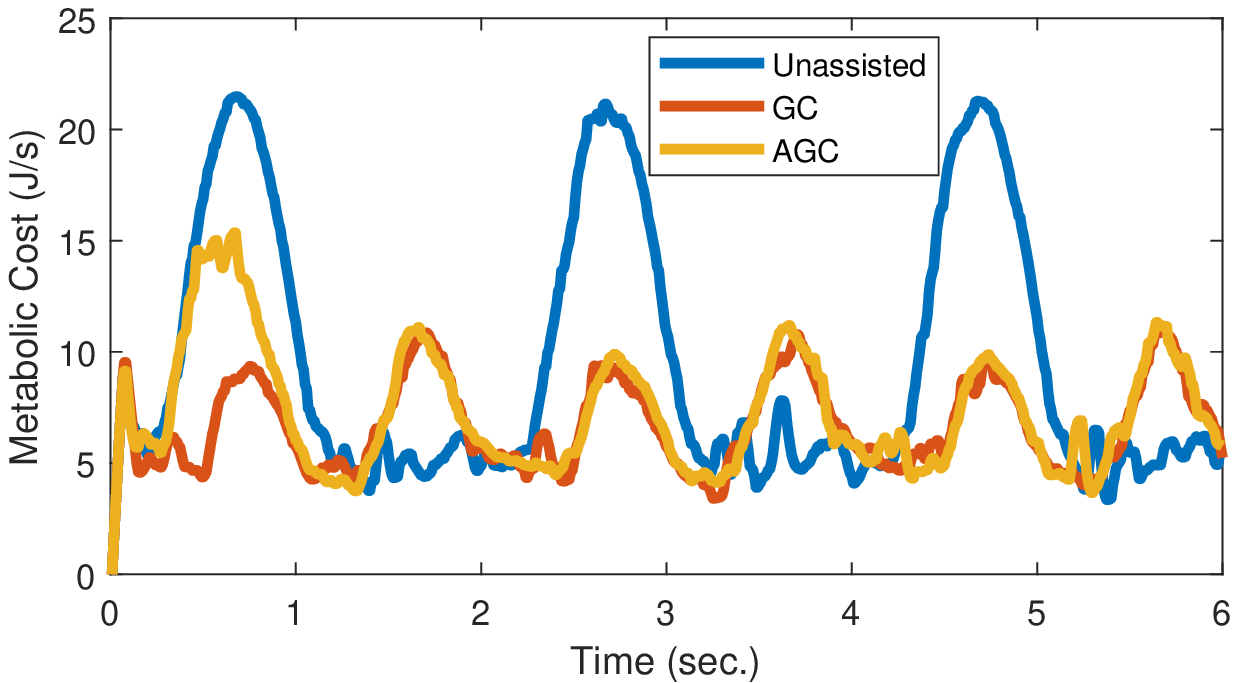}
     \caption{Metabolic Cost}
      \label{Metabolic_cost}  
	 \end{subfigure}  
	 \begin{subfigure}{1.1\columnwidth}
	 \flushleft
      \includegraphics[width = 3.6in]{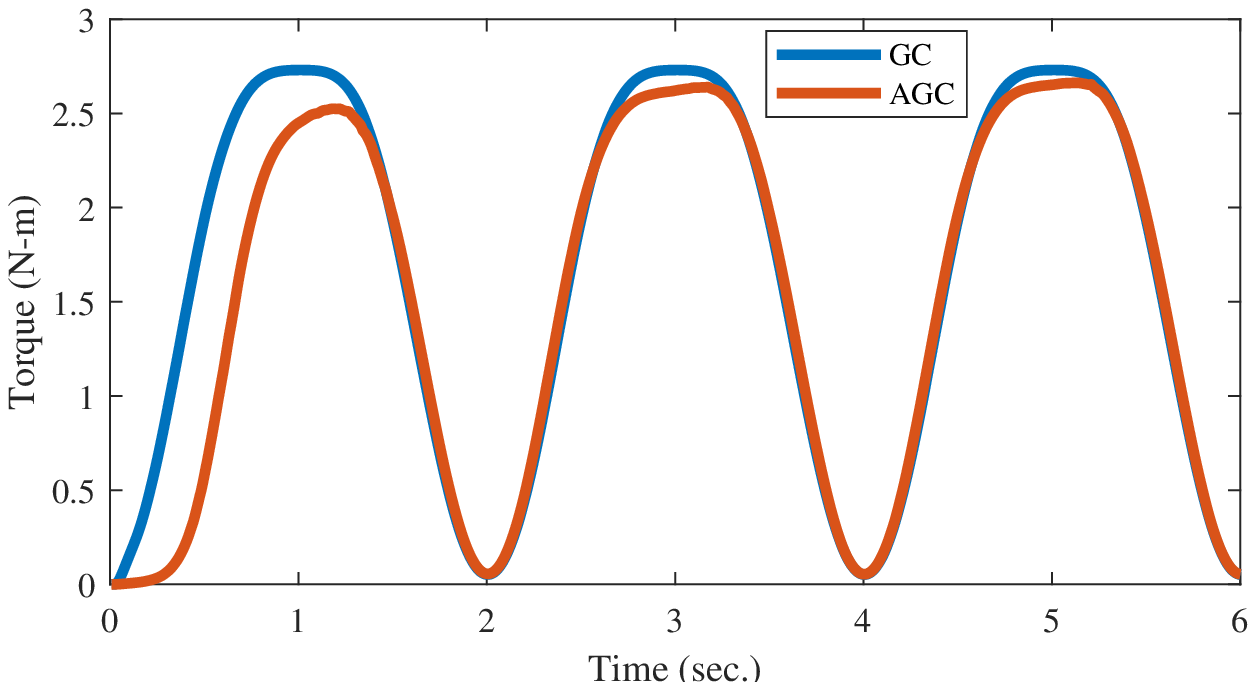}
      \caption{Assistive Torque $\tau_a(t)$ at the Elbow}
      \label{Assisted_torque}  
	  \end{subfigure}
      \begin{subfigure}{1.1\columnwidth}
	 \centering
      \includegraphics[width = 3.6in]{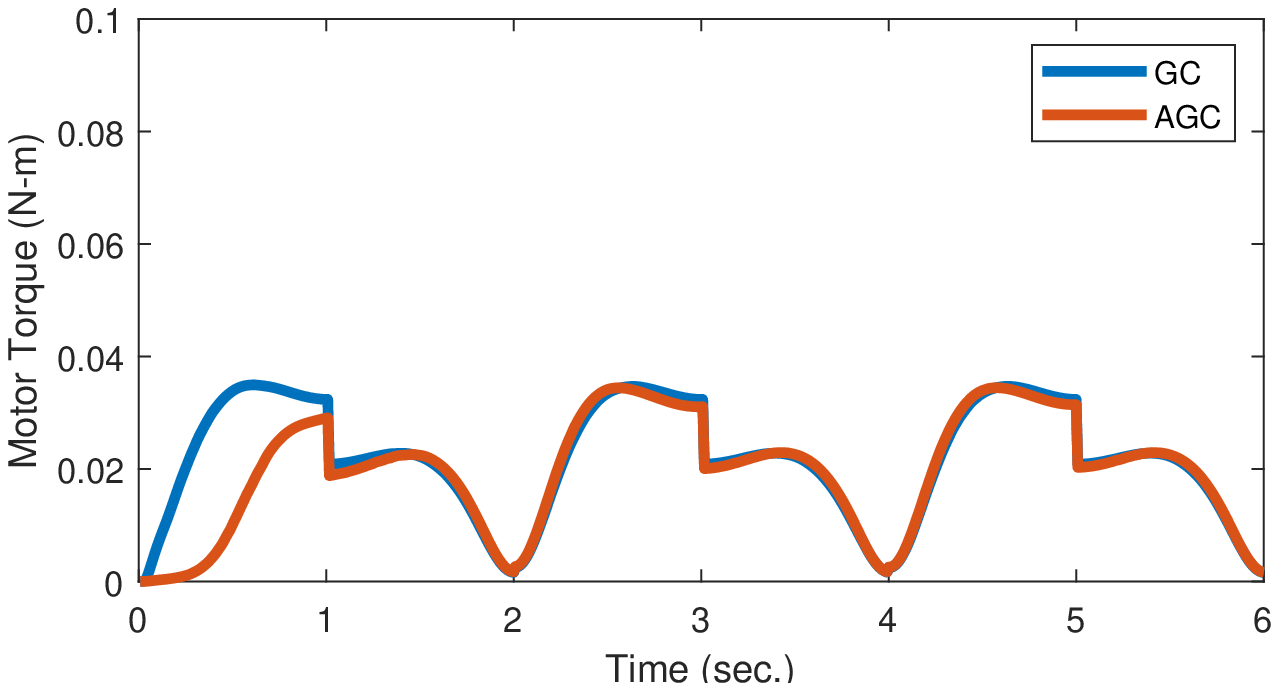}
      \caption{Motor Torque}

      \label{Motor Torque}  
	  \end{subfigure}
     \begin{subfigure}{1.1\columnwidth}
	 \centering
      \includegraphics[width = 3.6in]{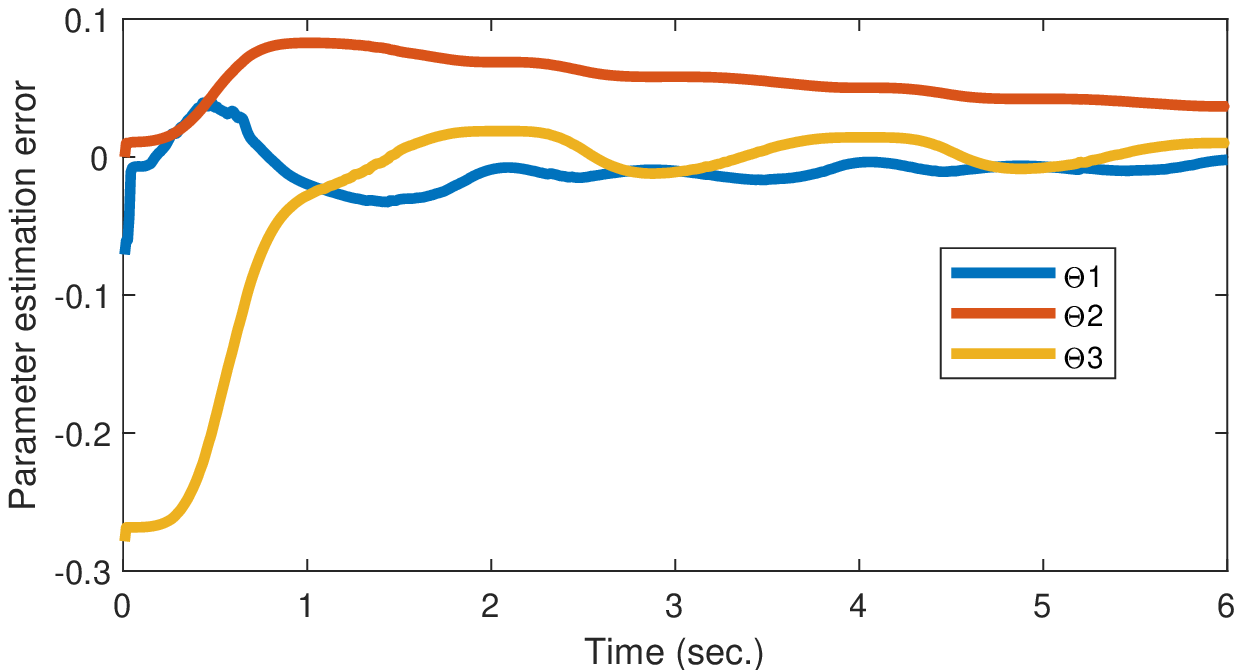}
      \caption{Parameter Estimation Error with AGC}
      \label{Parameter_error}  
	  \end{subfigure}  
	 \caption{ Simulation results: no external load. (a) Elbow joint trajectory under AGC assistance, (b) human elbow torque comparison (Unassisted, GC, AGC), (c)  metabolic cost comparison (GC vs AGC), (d)  assistive torque comparison (GC vs AGC), (e)  motor torque comparison (GC vs AGC) and (f) AGC assistance: parameter estimation error.  }
\label{fig:6}
\end{figure*}

A sinusoidal reference trajectory $\theta^r = \theta_m \sin{(\omega_h t)}$ has been designed for simulating the elbow flexion and extension with a simulation time of $t = 4~sec$ where $\theta_m = 90~\deg$ and $\omega_h = \pi/2~rad/s$. The choice of these parameters is based on the fact that generally, humans flex their elbows about $90 \deg$ while lifting a heavy load, and the physiological frequency of motion for the human hand is in the range of $1.2 - 1.5~rad/sec$. As mentioned previously, the GC implementation requires all the system parameters, mentioned in Table \ref{tab:table2}, to be known. However, most system parameters including masses, lengths, damping constant, and inertia are considered  unknown for the AGC control framework. The initial estimates of the controller parameters for the AGC law are given in Table \ref{tab:table3}.

\begin{table}[h!]
\centering
\caption{Controller Parameters}
   \label{tab:table3}
    \begin{tabular}{ |c|c| } 
    \hline
    Parameter  & Numerical (\%)\\
    \hline
    $\Theta_{1}(0)$ & $0.1$  \\
    \hline
    $\Theta_{2}(0)$ & $0.1$  \\
    \hline
    $\Theta_{3}(0)$ & $0.1$  \\
    \hline
    $\Lambda$ & $diag(0.05)$  \\
    \hline
 
\hline
 \end{tabular}
\end{table}

In this study, we evaluated the performance of the proposed methodology under varying payload conditions (0kg, 3kg, 5kg, 10kg). In the first case, no external load is considered. The reference and actual trajectory of the elbow joint angular position with AGC assistance are shown in Figure \ref{traj_1}. The torque provided by the human on the elbow joint for the unassisted, GC assist, and AGC assist scenarios are depicted in Figure \ref{Human_torque} and show that the exosuit (GC and AGC assist) leads to a reduction in human effort over time. Additionally, the human torque in the AGC method is observed to converge to the GC assist torque. This conclusion is further supported by the comparison of the torque provided by the cable-driven actuator for both the GC and AGC control methods, as shown in Figure \ref{Assisted_torque}. The motor torque computed using the power transmission coefficient $G$ in \eqref{G_transmission}, for both GC and AGC strategies, is shown in Figure \ref{Motor Torque}.

The proposed prediction error-based adaptive framework is also seen to be effective in learning the unknown system parameters, as seen from the parameter estimation error plots in Figure \ref{Parameter_error}. The estimation error for the third parameter, denoting the gravity term, is seen to be converging to zero, further highlighting the efficacy of the proposed framework. 

\begin{figure}[h!]
\centering
\includegraphics[width=0.95\linewidth]{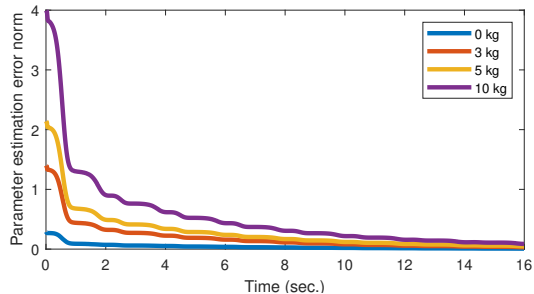}
\caption{Parameter Estimation Error $\|\tilde{\boldsymbol{\Theta}}(t)\|_2$ with varying payload}
\label{fig:7}
\end{figure}

The results of trajectory tracking, human and actuator torques, and parameter estimation error for different payloads (3kg, 5kg, 10kg) follow a similar trend to the no-load case. Figure \ref{fig:7} illustrates the 2-norm of the parameter estimation error for different payloads. 

\subsection{Metabolic cost comparison}

\begin{figure}[h!]
\centering
\includegraphics[width=0.95\linewidth]{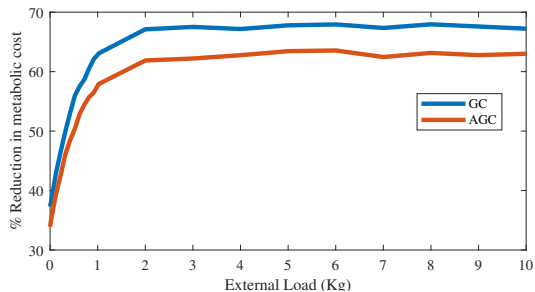}
\caption{\% Reduction in Metabolic cost when using GC and AGC compared to the unassisted case}
\label{fig:8}
\end{figure}

The results of the percentage reduction in metabolic cost with GC and AGC are depicted in Figure \ref{fig:8}. Here, we observe that the decrease in metabolic cost initially increases with increasing load and eventually stabilizes at 67\% for the GC method and 63\% for the AGC method. The difference in the reduction of metabolic cost between the GC and AGC methods is due to AGC's initial transient learning phase, beyond which both GC and AGC have comparable performance. 

\begin{figure*}[h!] 
	  \begin{subfigure}{0.81\columnwidth}
	  \flushleft
      \includegraphics[width = 3.6in]{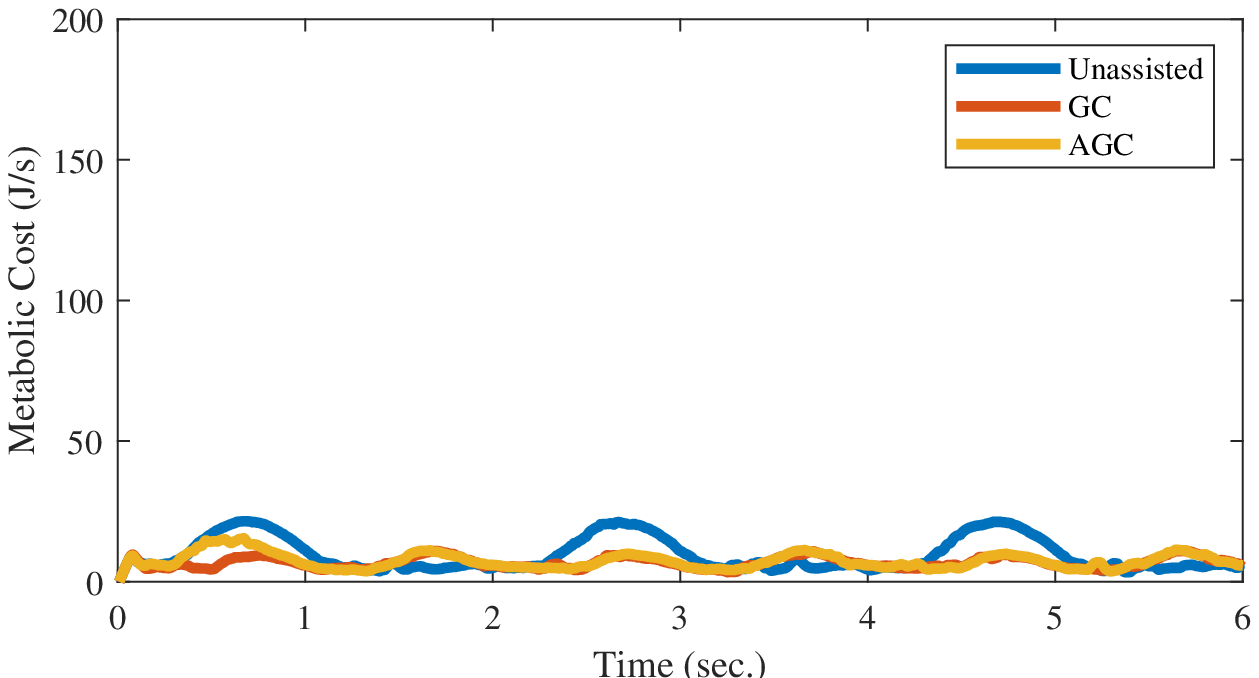}
      \caption{No load}
      \label{state1}  
	  \end{subfigure}  \begin{subfigure}{1.1\columnwidth}
	  \flushleft
      \includegraphics[width = 3.6in]{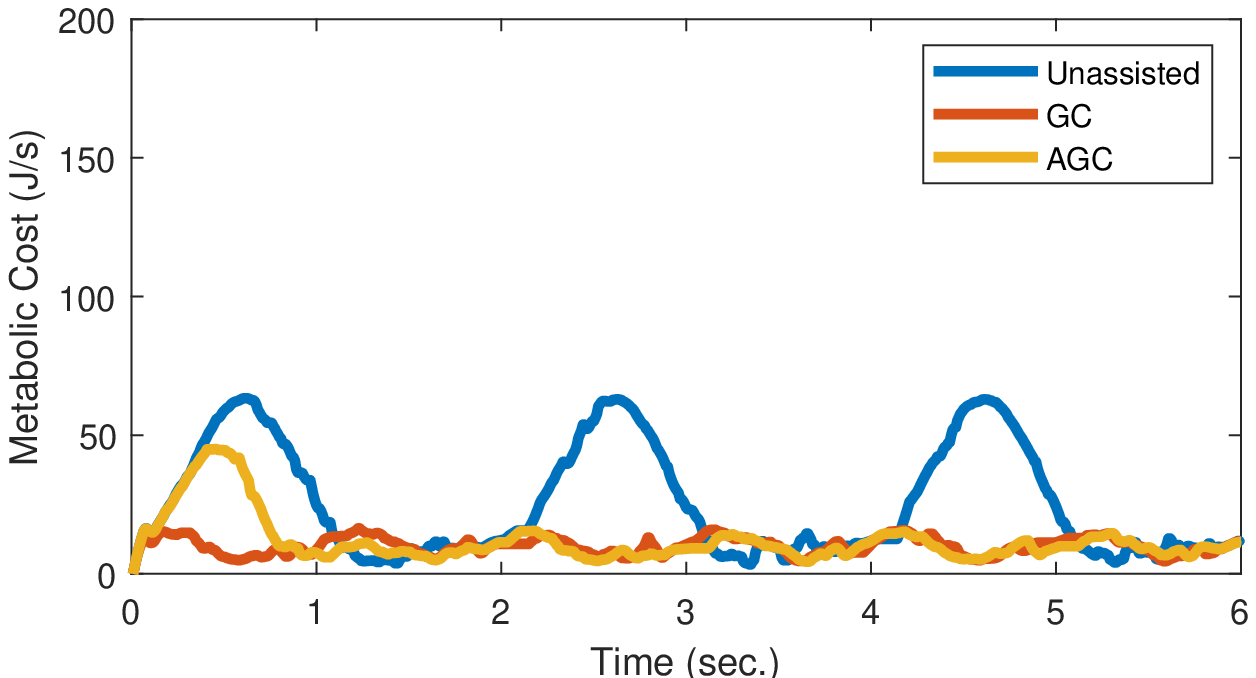}
      \caption{3 kg load}
      \label{state2}  
	 \end{subfigure}  
	 \begin{subfigure}{0.9\columnwidth}
	 \flushleft
      \includegraphics[width = 3.6in]{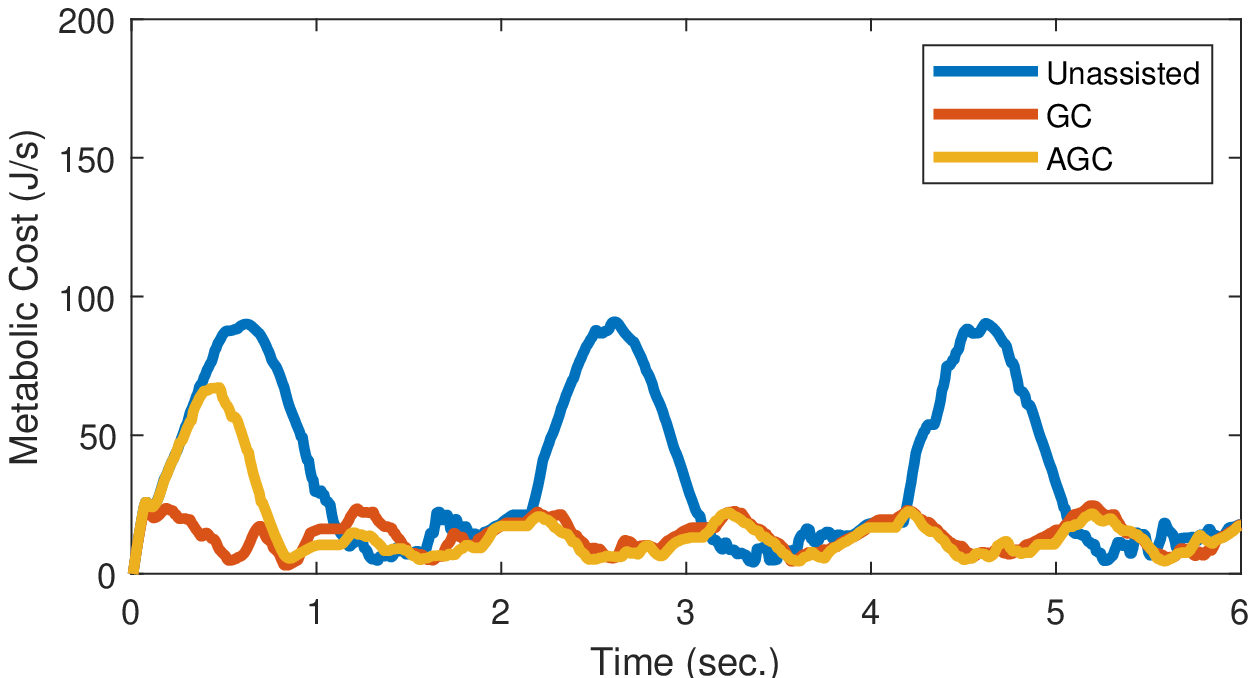}
      \caption{5 kg load}
      \label{state3}  
	  \end{subfigure}
	 \begin{subfigure}{1.1\columnwidth}
	 \centering
      \includegraphics[width = 3.6in]{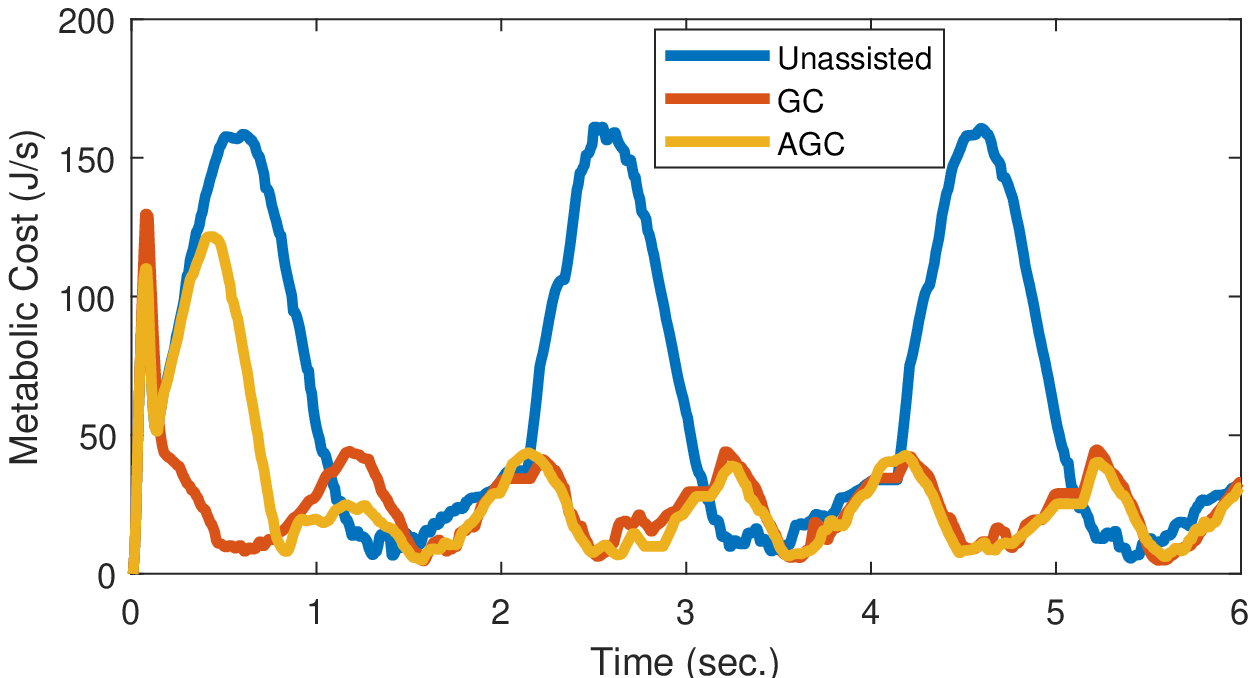}
      \caption{10 kg load}
      \label{state4}  
	  \end{subfigure}  
	 \caption{ The metabolic cost comparison plots (Unassisted, GC, AGC) with varying payloads.  }
\label{fig:9}
\end{figure*}

The results for reduction in metabolic cost for various load conditions (no-load, 3 kg, 5 kg, and 10 kg) for both elbow flexion and extension with and without external assistance are shown in Figure \ref{fig:9} (a-d), where we can see that after the initial learning phase, the AGC method begins to perform at par with the GC method.

\subsection{Discussions}
The quantitative effect of the control algorithm on the reduction of metabolic cost in a dynamic simulation can also be observed through the reduction in the RMS (root mean squared) value of the physiological torque provided by the human hand.
For a sinusoidal trajectory from extended to flexed elbow position and vice-versa, the GC and AGC approach results in a reduction of the hand torque (RMS) as shown in Table \ref{tab:table5}.\\

\begin{table}[ht]
\centering
\caption{Percentage reduction in the human muscle torque}
\label{tab:table5}
   \begin{tabular}{ |c|c|c| } 
    \hline
    Additional Payload & \multicolumn{2}{c|}{Reduction in Elbow Torque (\%) } \\
    \cline{2-3}
     (Kg)& GC assistance  & AGC assistance \\
    \hline
    0 &  35 & 28  \\
    \hline
    3 &  65 & 50 \\
    \hline
    5 & 67 & 66 \\
    \hline
    10 &  67 & 66  \\

\hline
 \end{tabular}
\end{table}

The torque and metabolic cost plots suggest that the application of an external actuator in a soft exosuit can suitably provide augmentation to the elbow joint. However, these results have to be validated with an experimental setup.

\section{Conclusion} \label{sec5}
A self-tuning adaptive gravity compensation control law has been designed in this work to reduce the human effort needed to track a desired elbow trajectory with an unknown load. The mathematical expressions of the control law along with the adaptation law are derived from the Lyapunov-based stability analysis which has been verified through simulation studies in different payload scenarios. The advantage of these control approaches (GC and AGC) over conventional feedback-based (e.g. PID) approaches is that the controller does not need information about the desired trajectory which is difficult to obtain in real-time. Further, unlike GC, the AGC method obviates the requirement of system parameters using an online parameter estimator.The simulation results show that with online parameter learning, the performance of the AGC method, in terms of reduction of the human torque and metabolic cost, is comparable to that of the GC method after an initial transient learning period. Although the implementation of the GC control on a processor might be simple, the AGC control law requires more computation and sensing, i.e. acceleration and human torque estimation. Future work will involve the implementation of the proposed controller on an experimental soft exosuit hardware prototype.

\bibliographystyle{IEEEtran}
\bibliography{ref1}

\end{document}